\documentclass[journal,twoside,web]{ieeecolor}
\usepackage{generic}
\usepackage{cite}
\usepackage{graphicx}
\usepackage{hyperref}
\usepackage{textcomp}
\usepackage{amsmath,amssymb,amsfonts}
\usepackage{algorithmic,algorithm2e,float}
\SetKwInput{KwInput}{Input}                
\SetKwInput{KwOutput}{Output}              
\SetKwInput{KwRepeat}{Repeat}
\SetKwInput{KwUntil}{Until}
\SetKwInput{KwReturn}{Return}
\usepackage{amsfonts}
\usepackage{graphicx}
\usepackage{textcomp}
\usepackage{tabularx}
\usepackage{array}
\usepackage{booktabs}
\usepackage{threeparttable}  
\usepackage{multirow}
\usepackage{bm}
\usepackage{amsmath}
\usepackage{subfigure}
\usepackage{amssymb}
\usepackage{pifont}
\usepackage{adjustbox} %
\newcommand\dy[1]{{\color{black}#1}}

\def\BibTeX{{\rm B\kern-.05em{\sc i\kern-.025em b}\kern-.08em
    T\kern-.1667em\lower.7ex\hbox{E}\kern-.125emX}}
\begin{document}
\title{MASA-TCN: Multi-anchor Space-aware Temporal Convolutional Neural Networks for Continuous and Discrete EEG Emotion Recognition}
\author{Yi Ding, \IEEEmembership{Member, IEEE}, Su Zhang, Chuangao Tang, and Cuntai Guan \IEEEmembership{Fellow, IEEE}
\thanks{Cuntai Guan is the Corresponding Author. E-mail: ctguan@ntu.edu.sg}
\thanks{Yi Ding and Su Zhang contribute equally for this work.}
\thanks{Yi Ding, Su Zhang, and Cuntai Guan are with the School of Computer Science and Engineering, Nanyang Technological University, 50 Nanyang Avenue, Singapore, 639798.}
\thanks{Chuangao Tang is with the Key Laboratory of Child Development and Learning Science (Ministry of Education), School of Biological Science and Medical Engineering, Southeast University, Nanjing, 210096, China}
\thanks{This work was supported by the RIE2020 AME Programmatic Fund, Singapore (No. A20G8b0102).}
}
\maketitle
\begin{abstract}
Emotion recognition from electroencephalogram (EEG) signals is a critical domain in biomedical research with applications ranging from mental disorder regulation to human-computer interaction. In this paper, we address two fundamental aspects of EEG emotion recognition: continuous regression of emotional states and discrete classification of emotions. While classification methods have garnered significant attention, regression methods remain relatively under-explored. To bridge this gap, we introduce MASA-TCN, a novel unified model that leverages the spatial learning capabilities of Temporal Convolutional Networks (TCNs) for EEG emotion regression and classification tasks. The key innovation lies in the introduction of a space-aware temporal layer, which empowers TCN to capture spatial relationships among EEG electrodes, enhancing its ability to discern nuanced emotional states. Additionally, we design a multi-anchor block with attentive fusion, enabling the model to adaptively learn dynamic temporal dependencies within the EEG signals. Experiments on two publicly available datasets show that MASA-TCN achieves higher results than the state-of-the-art methods for both EEG emotion regression and classification tasks. The code is available at \textit{https://github.com/yi-ding-cs/MASA-TCN}
\end{abstract}


\section{Introduction}
\label{sec:introduction}

\IEEEPARstart{E}{motion} recognition, \dy{also known as emotional artificial intelligence \cite{7946165, Li_2023_CVPR}, leverages machine learning to understand human emotions, crucial for addressing emotion-related mental disorders like anxiety, depression, and autism spectrum disorder (ASD). It employs both categorical and dimensional models, with the valence-arousal-dominance (VAD) model \cite{7946165} being prominent for evaluating emotions across valence (negative to positive), arousal (passive to active), and dominance (emotion strength) dimensions. Unlike affective signals from physiological responses, speech, and facial expressions \cite{Li_2019_CVPR}, electroencephalogram (EEG) is notably effective in emotion recognition. As a cost-efficient, non-invasive, and user-friendly brain imaging tool, EEG uniquely captures the brain's inherent emotional neural activities, distinguishing it from other modalities that subjects could easily conceal.}

\dy{We differentiate between discrete emotional state classification (DEC) and continuous emotion regression (CER), two key tasks in EEG emotion recognition, covering aspects such as data acquisition, annotation, representation learning, and evaluation. Both tasks utilize a similar data acquisition approach, where subjects watch film clips to capture facial expressions, EEG signals, and other emotional cues \cite{5871728, 5975141, HSU2022118873}. The distinction between DEC and CER lies in annotation, representation learning, and evaluation. DEC assigns a single discrete label per trial \cite{9762054}, while CER maps each trial with a series of values indicating emotional fluctuations \cite{ZHANG2022108833}. In representation learning, DEC may underemphasize long-term dynamics, crucial for CER's sequence prediction. For evaluation, DEC relies on accuracy and F1 score \cite{9762054}, whereas CER uses consistency metrics like RMSE, PCC, and CCC \cite{ZHANG2022108833}. Both tasks face challenges, especially in generalized settings where classifiers are tested on unseen data \cite{9762054}.}

\dy{Those challenges have caught the interest of many researchers in recent years \cite{7946165}, especially for DEC tasks. In DEC tasks, traditionally, different types of features are extracted from pre-processed EEG signals \cite{7938737, 8634938}. Then the shallow learning methods were applied, such as support vector machine (SVM). With the rapid development of deep learning in domains such as computer vision \cite{he2016deep,dosovitskiy2021an}, natural language processing (NLP) \cite{kenton2019bert,NEURIPS2020_1457c0d6,ding-etal-2022-globalwoz, ding-etal-2023-gpt}, and graphs \cite{luo2024learning,kipf2017semisupervised,10366850}, more and more researchers apply different types of neural networks to the BCI domain \cite{https://doi.org/10.1002/hbm.23730, 8310961, Lawhern_2018,CHEN2023521, 8897723, 10025569, doi:10.1080/27706710.2023.2181102}. Among deep learning methods, there are two types of learning paradigms. The first one feeds hand-crafted features to neural networks, extracting the spatial-temporal patterns of the features via different types of neural networks, namely, convolutional neural networks (CNN) \cite{JIAO2018582}, graph convolutional neural networks (GCN) \cite{8320798, 10216317, 10340644}, recurrent neural networks (RNN) \cite{8275511}, LSTM \cite{8736804}, transformers \cite{10345766, 9684393, 9857970, 10209178}, and other hyper networks \cite{8621147, 9413635, 10.3389/fnbot.2019.00037, 9963543}.} With the automatic feature-learning ability of CNN, using EEG signals directly becomes a new trend \cite{9762054, 9204431}. Compared to the well-studied DEC task, the CER task is less explored, with fewer databases and methods. Among existing studies, the majority use visual and audio modalities \cite{9607640, 10.1145/3347320.3357688}. There are only a few works \cite{ZHANG2022108833, 7112127} that focus on the EEG-based CER problem. LSTM was utilized to do the regression of continuous emotional labels using PSD features \cite{7112127}. TCN  was further explored to learn from relative PSD (rPSD) features of EEG, achieving the state-of-the-art (SOTA) results for CER task using EEG \cite{ZHANG2022108833}. However, those two methods all utilized flattened feature vectors. Hence, the spatial information of EEG signals were not capably learned by the neural networks. Since the label is continuous in time, learning temporal dynamic patterns are essential which can be addressed well by using TCN. 

A natural question is: \textit{How can we empower the TCN with spatial learning abilities to further improve regression performance?} Furthermore, \textit{can we propose a unified model that can handle both regression and classification well?} To answer the above-mentioned questions, we designed a novel multi-anchor space-aware TCN (MASA-TCN) as a unified model for both DEC and CER tasks. A space-aware temporal layer (SAT) is designed to give TCN the ability to learn spatial-spectral patterns from rPSD. The SAT layer has two types of kernels: spectral context kernels as well as spatial fusion kernels. Spectral context kernels extract different spectral patterns channel by channel. The spatial fusion kernels serve as the spatial pattern learners that extract the patterns among different channels. Besides, a multi-anchor attentive fusion block (MAAF) is proposed to extract the dynamic temporal patterns. It parallelly applies different-length causal temporal kernels. Then the outputs are attentively fused. Two publicly available datasets,  MAHNOB-HCI \cite{5975141} and DEAP \cite{5871728}, were utilized to evaluate the proposed MASA-TCN for CER and DEC tasks. MASA-TCN was compared with the SOTA methods for CER and DEC tasks. Based on the experiment results, MASA-TCN achieved better regression and classification results and set new SOTA results for those two tasks. Extensive experiments and visualizations were conducted to better understand the proposed method. The results suggest that enabling TCN to extract spatial patterns improves its performance, and the network width plays a more important role for MAS-TCN than the depth does. The experiments also suggest that attentive fusion and early spatial fusion are important for performance improvements on CER tasks.

We summarize the contributions of this work as: 
\begin{itemize}
\item We developed MASA-TCN, a novel unified model for both EEG emotion regression and classification tasks. 
\item The space-aware temporal layer was designed to enable TCN to extract spatial-spectral patterns.  
\item Additionally, we proposed a multi-anchor attentive fusion block to capture temporal dynamic patterns.
\item Extensive analyses and ablation studies were conducted to evaluate the importance of each module in MASA-TCN.
\end{itemize}

The subsequent sections of this article are structured as follows: Section 2 introduces some preliminary concepts. In Section 3, we delve into the intricate details of MASA-TCN. Section 4 provides an overview of the datasets and experiment settings. Section 5 presents the results and analysis. Lastly, we offer a comprehensive discussion and conclusion in Sections 6 and 7, respectively.

\begin{figure*}[htp]
    \centering
    \includegraphics[width=\linewidth]{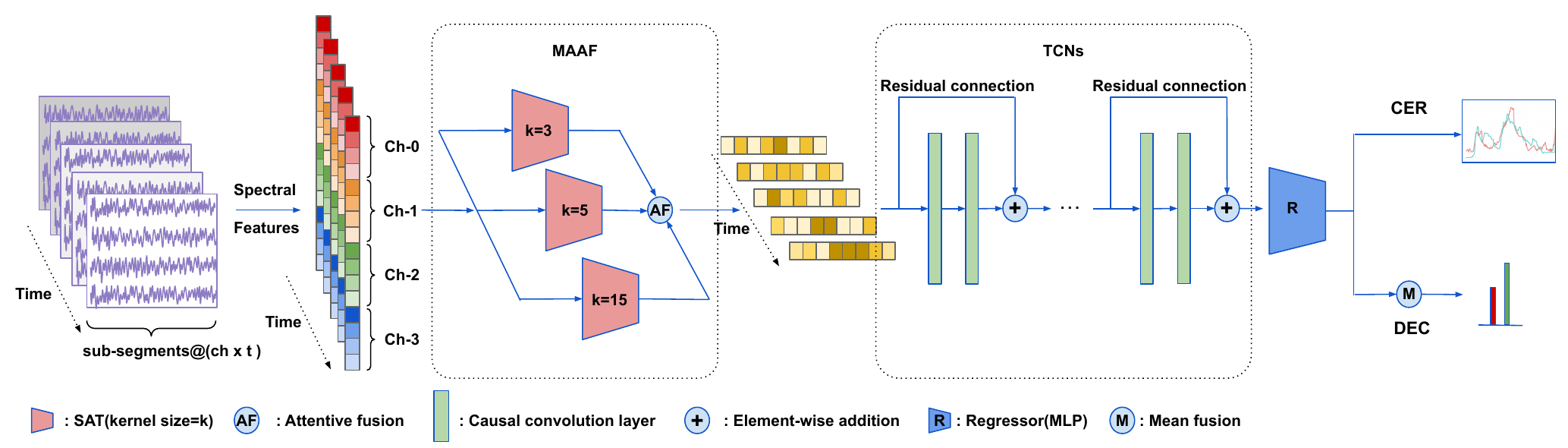}
    \caption{The architecture of our MASA-TCN. There are four main parts of MASA-TCN: feature extraction block, MAAF block, TCN block, and regression/classification block. A sequence of five four-EEG-channel sub-segments is utilized as an example. The k of SAT refers to the length of the kernel in the temporal dimension. Best viewed in color.}
    \label{fig:MASA}
\end{figure*}

\begin{figure}[ht]
    \centering
    \includegraphics[width= 9cm]{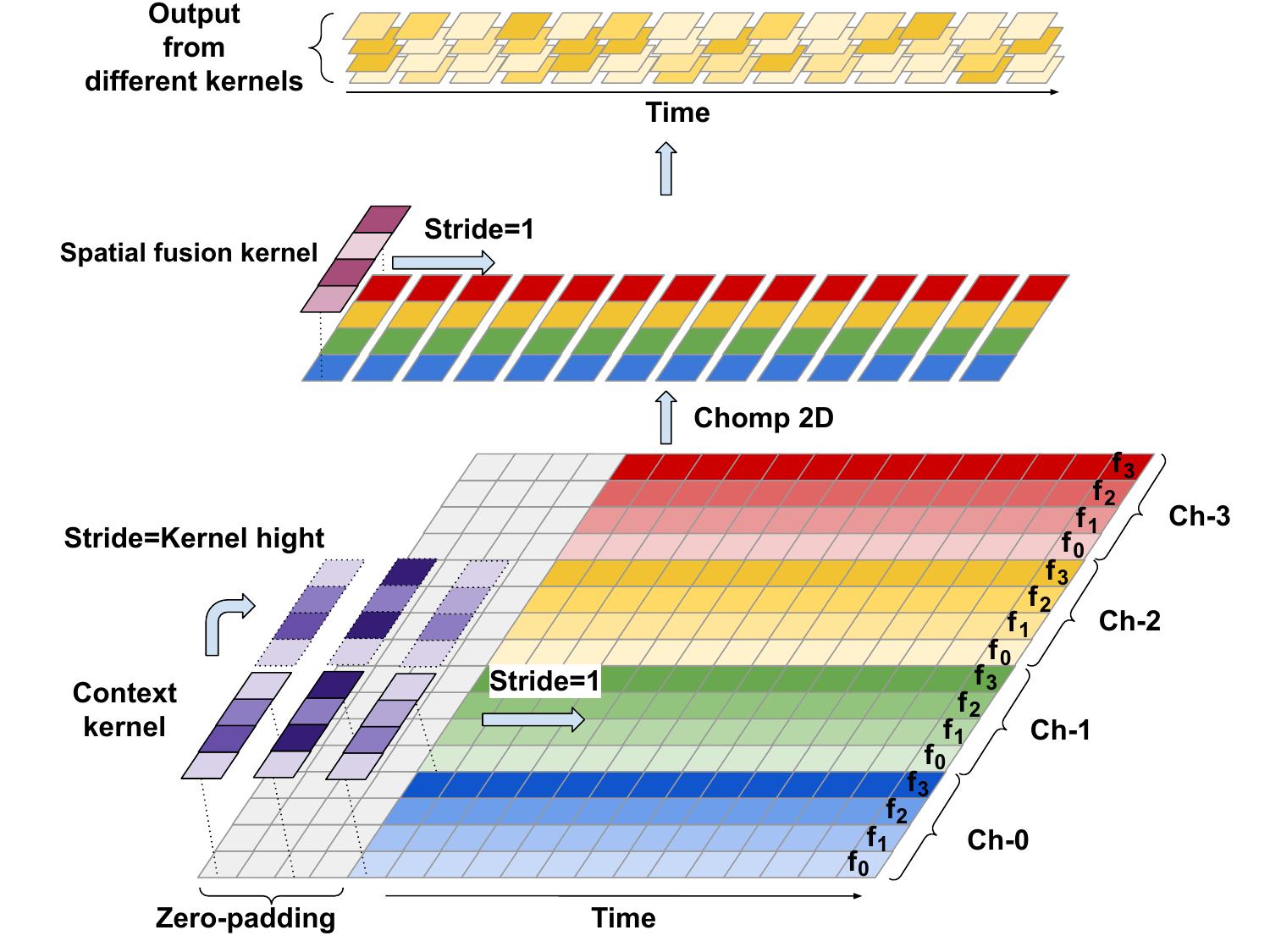}
    \caption{Space-aware temporal convolutional layer. The SAT has two types of convolutional kernels: context kernels that extract the spectral patterns channel by channel and spatial fusion kernels that learn spatial patterns across all the channels. A context kernel of size (4, 3) is utilized for example. And a four-EEG-channel sample with four spectral features in each EEG channel is used as the example. Zero padding is added to make the context kernel a causal kernel along temporal dimension. Only one kernel for each type of CNN kernels is shown in the diagram for better view, the final output (on the top) consists of the outputs from more kernels (4 is utilized as the number of kernels of each type for demonstration purpose). \dy{Ch-$n$ represents $n$-th EEG channel, and $f_{n}$ is the $n$-th frequency band. }Best viewed in color.}
    \label{fig:SAT}
\end{figure} 

\section{Preliminaries} 
\subsection{Problem formulation}
There are two types of EEG emotion recognition tasks to be addressed in this paper: CER and DEC. We provide a more formal description on the data annotation of CER and DEC. Given $N$ trials of continuous EEG signal $[\textbf{X}_{0}, ... , \textbf{X}_{N-1}], \textbf{X} \in \mathbb{R}^{C \times T}$, where $C$ presents EEG electrode numbers, $T$ is the number of temporal data points. Typically, the entire trial is divided into shorter segments, denoted by $\Bar{\textbf{X}}_{i}, i \in [0, 1, ..., n-1]$, using a sliding window with or without overlap to train the neural networks. For CER, the labels are $\textbf{y}_{CER}=[\textbf{y}_{0}, ... , \textbf{y}_{n-1}], \textbf{y} \in \mathbb{R}^{1 \times T/f_{s}^{y}}$, where $f_{s}^{y}$ is the sampling rate of the continues labels. Because the label of each trial in CER is continuous in the temporal dimension, the label is also cut into shorter segments as is done for the EEG data. The target of CER is to learn $f (\Theta):\textbf{X}_{i} \to \textbf{y}_{CER}$, which can:
\begin{equation}
    \underset{\Theta}{argmin}\sum_{n-1}^{i=0}\Psi(f(\textbf{X}_{i}), \textbf{y}_{CER, i}),
\end{equation}
where $\Theta$ is the trainable parameters of $f(\cdot)$ and $\Psi(\cdot)$ is the regression loss. 

For DEC, the labels are $y_{DEC}=[y_{0}, ... , y_{n-1}], y \in \mathbb{R}$. Because there is one label for each trial in DEC, all the segments within one trial share the same label. The target of DEC is to learn $f (\Theta):\textbf{X}_{i} \to y_{DEC}$, which can:
\begin{equation}
    \underset{\Theta}{argmin}\sum_{n-1}^{i=0}\Upsilon(f(\textbf{X}_{i}), y_{DEC, i}),
\end{equation}
where $\Theta$ is the trainable parameters of $f(\cdot)$ and $\Upsilon(\cdot)$ is the cross-entropy loss. 

\subsection{Neural networks for temporal pattern recognition}
\dy{This section introduces two neural networks for temporal pattern recognition: RNN and TCN. RNNs, distinct from feed-forward networks, leverage previous outputs as inputs, incorporating internal states to learn temporal dynamics. LSTM \cite{6795963}, a variant of RNN, efficiently models sequential patterns using a cell state for information retention and gates for regulating data updates. Bidirectional LSTMs enhance pattern learning by processing sequences in reverse order. Gated recurrent units (GRU) \cite{https://doi.org/10.48550/arxiv.1409.1259}, a simpler LSTM alternative, achieves comparable performance with fewer gates. LSTM's capability in temporal pattern extraction from flattened PSD vectors for CER was demonstrated in \cite{7112127}. TCN, introduced for action segmentation \cite{8099596}, employs causal and dilated convolutions for sequential modeling, with residual connections enhancing its performance \cite{https://doi.org/10.48550/arxiv.1803.01271}. Zhang et al. \cite{ZHANG2022108833} applied TCN with rPSD features for CER, surpassing LSTM's results \cite{7112127}.

However, the spatial relations among electrodes remain underexplored with both LSTM and TCN relying on flattened spectral features. To address this, we introduce a space-aware temporal convolutional layer for TCN to effectively learn spatial patterns in CER tasks.} 
\dy{Considering emotion's temporal variability \cite{doi:10.1177/1754073915590618, 9698041}, a multi-anchor attentive fusion block is devised to enhance temporal dynamics modeling in affective EEG data. Unlike previous approaches using varying dilation rates, we employ 1D causal convolutional kernels of different lengths to capture the dynamic temporal dependencies in emotional processes. In addition to CER, which necessitates the model's ability to capture temporally continuous changes, this design can also be applied to DEC. DEC demands one overall prediction per input, which can be achieved by employing a mean fusion strategy across predictions from all sub-segments within each input segment.}

\section{Method}
In this section, the detailed introduction of each functional component in \dy{MASA-TCN} is presented. TCN has superior sequential pattern modeling ability. However, EEG data has spatial and temporal patterns to be extracted for the regression and classification tasks. Previous works use flattened rPSD features as the input to TCN directly, which can not extract the spatial pattern effectively. A space-aware temporal convolutional layer (SAT) is proposed to extract spatial-spectral patterns of EEG using TCN. Besides, to better learn the temporal dynamics underlying emotional cognitive processes that might appear in different time scales, we design a \dy{MAAF}. The MAAF consists of three parallel SATs with different lengths of 1D causal convolution kernels. The outputs of these parallel SATs are attentively fused as the input to several TCN layers which learns the higher-level temporal patterns and generates the final hidden embedding. For the CER tasks, a linear layer is utilized as a regressor to map the hidden embedding to the continuous labels. For the DEC tasks, because these segments in time order share one label of that trial, a sum fusion layer is utilized to generate the final output instead of using a linear layer to get a single output. The structure of MASA-TCN is demonstrated in Fig.~\ref{fig:MASA}.

\subsection{Input construction}
The construction of the network input is illustrated first to better understand the algorithm. As mentioned in Section 2.1, the EEG data of each trial is cut into shorter segments. Note that the sampling rates of the EEG data and continuous label are different, the former is much higher than the latter, e.g. 256Hz vs 4Hz in MAHNOB-HCI. Then the segments are further segmented into sub-segments along temporal dimension. Sliding windows with overlaps are applied to make sure sub-segments are synchronized to each value of the continuous label for CER. For each sub-segment, it is still a 2D matrix, which has spatial and temporal dimensions. Because the sub-segments are in time orders, they can be regarded as frames in a video. For each sub-segments, averaged rPSDs in 6 frequency bands are calculated as described in \cite{ZHANG2022108833}. We flatten the rPSDs along the EEG channel dimension, resulting in a feature vector:
\begin{equation}
    \textbf{v} = [[p_{c_{0}}^{f_{0}},..., p_{c_{0}}^{f_{F-1}}], ... , [p_{c_{C-1}}^{f_{0}},..., p_{c_{C-1}}^{f_{F-1}}]]\label{eq:input-vector},
\end{equation}
where $p $ is the averaged rPSD, $C$ equals to the EEG channel number, $F$ equals to the total number of the frequency bands, and $[\cdot]$ denotes the concatenation. \dy{$p_{c_{0}}^{f_{0}}$ represents the averaged rPSD of the channel $c_{0}$ in frequency band $f_{0}$.} Hence, one input to the neural networks would be:
\begin{equation}
    \textbf{x} = [\textbf{v}_{0}, ..., \textbf{v}_{t-1}]^{T}\label{eq:input-sample},
\end{equation}
where t represents the total number of the rPSD vectors within one segment. 

TCN utilizes 1D CNN along the temporal dimension and treats the feature vector that contains spectral and spatial patterns as the channel dimension of 1D CNN. Hence, the spectral patterns across EEG channels as well as the spatial relations among EEG electrodes are not capably learned. Instead of treating the feature vector dimension as the channel dimension of 1D CNN, we treat the input to TCN as a 2D matrix, whose dimensions are feature and time.

\subsection{Space-aware temporal convolutional layer}
The SAT has two types of convolutional kernels: context kernels that extract the spectral patterns channel by channel and spatial fusion kernels that learn spatial patterns across all the channels. The structure of SAT is shown in Fig. \ref{fig:SAT}. 

Given the input $\textbf{x}=[\textbf{v}_{0}, ... , \textbf{v}_{t-1}]^{T}, \textbf{v} \in \mathbb{R}^{1 \times C*f}$ introduced in Section 3.1, the first type of the CNN kernels in SAT is the 2D causal convolutional kernel whose size, step, and dilation are $(f, k), (f, 1)$, and $(1, 2)$, where $f$ is the number of frequency bands used to calculate rPSDs and $k$ is the length of the CNN kernel in temporal dimension. Note that the default dilation step is 1 instead of 0 in PyTorch \cite{NEURIPS2019_9015} library, which means there is no dilation in that dimension if the dilation step is set as 1. Because the step in the feature dimension is the same as the height of the kernel, it can learn spectral contextual patterns across EEG channels. Hence, it is named the context kernel. The context kernel can learn spectral patterns as well as temporal dynamics at the same time due to its 2D shape. Different from WaveNet \cite{45774} that has dilation steps of $1, 2, 4, ... 2^{n-1}$, where n is the number of layers, the first layer of MASA-TCN has a dilation of 2 in the temporal dimension. There are two reasons. The first reason is that the higher dilation step helps to get more discriminative information, considering the fact that the rPSD is calculated using overlapped sliding windows so that the adjacent vectors are highly correlated. The second reason is that discarding the TCN layer with a dilation step of 1 leads to smaller model size, without any compromises on the receptive field. Due to the causal convolution, the temporal dimension of the input and output are the same. Hence, we can get the output $\textbf{H}_{context} \in \mathbb{R}^{s, C, t}$, where $s$ is the number of context kernels, $C$ is the number of EEG channels, and $t$ is the total number of the rPSD vectors within one segment. $\textbf{H}_{context}$ can be calculated by:
\begin{equation}
\begin{aligned}
    \textbf{H}_{context} = &\textrm{Conv2D}(\\
    &input=\textbf{x}, \\
    &kernel\_size=(f, k), \\
    &strides=(f, 1), \\
    &dilation=(1, 2)),\label{eq:h-context}
\end{aligned}
\end{equation}
\dy{where $Conv2D$ represents the 2D convolution applied to the input \textbf{x}, with $kernel_size$, $strides$, and $dilation$ as the parameters for the CNN operation. Note that these parameters are set to their default values in the PyTorch library unless specified otherwise. Given that $f$ denotes the number of frequency bands used for rPSD (refer to Section~\ref{sec_preproces}), $f_{\textrm{MAHNOB-HCI}}$ is 6 and $f_{\textrm{DEAP}}$ is 5. We set $k$ to values in the set [3, 5, 15], and an analysis to evaluate the effects of $k$ is conducted in Section~\ref{sec_kernel_size}.}

The output of the context kernels is spatially fused by spatial fusion kernels to learn the spatial patterns of EEG channels. The size, stride, and dilation of the spatial kernels are $(C, 1), (1, 1), (1, 1)$, respectively. This is the same as the commonly used spatial kernels of CNNs in BCI domains \cite{https://doi.org/10.1002/hbm.23730, Lawhern_2018}. Besides, it can be treated as an attentive fusion of all the EEG channels, with the weights of the 1D CNN kernel being the attention scores. After $s$ spatial fusion kernels, the size of the hidden embedding $\textbf{H}_{SF}$ becomes $(s \times 1 \times t)$. This process can be described as:

\begin{equation}
    \textbf{H}_{SF} = \textrm{Conv2D}(\textbf{x}, kernel\_size=(C,1)),\label{eq:h-SF}
\end{equation}
where the default values of strides $(1, 1)$ and dilation $(1, 1)$ are utilized. 
\subsection{Multi-anchor attentive fusion block}
There are two steps in the MAAF: 1) parallel SATs with different temporal kernel lengths and 2) attentive fusion of the output from these SATs. The architecture of MAAF is shown in Fig. \ref{fig:MASA}. TSception \cite{9762054} utilizes multi-scale temporal convolutional kernels to capture temporal dynamics that might happen at different time scales. Emotion varies from time to time, especially over longer duration \cite{9698041}. The duration of emotions varies from a few seconds to several hours \cite{doi:10.1177/1754073915590618}. 

Three parallel SATs with different temporal kernel sizes are utilized to capture those temporal dynamics in different time scales. In this paper, the temporal lengths of the context kernels are set to $k = [3, 5, 15]$, respectively. The longer the temporal length, the larger the temporal receptive field. Because the weights of the context kernels are distributed along the time dimension with the help of dilation steps, each weight is like an anchor on the time axis. Hence, we name these parallel SATs multi-anchor SATs. Besides, from a causal dependence perspective, different temporal kernel sizes may incorporate various previous outcomes to determine the subsequent output. We hypothesize that it can increase the robustness of the causal dependence in the temporal dimension underlying the continuous emotional cognitive process. The results in ablation studies also support the effectiveness of the multi-anchor design. The multi-anchor SATs can be described as:

\begin{equation}
    \textbf{H}_{MA}^{i} = \textrm{SAT}(\textbf{x}, kernel=(f, k_{i})), i \in [0, 1, 2],
\end{equation}
where SAT contains the sequential operation of Eq. \ref{eq:h-context} and Eq. \ref{eq:h-SF}. 

\dy{Different from TSception that directly concatenates the output of different scale kernels, an attentive fusion operation is adopted to combine the output from different SATs. First, the three outputs are concatenated along the kernel dimension (channel dimension of CNNs). Given $\textbf{H}_{MA}^{i} \in \mathbb{R}^{s \times 1 \times t}$, the concatenated output would be $\textbf{H}_{MA}^{cat} \in \mathbb{R}^{3*s \times 1 \times t}$. Then, a one-by-one convolutional layer with $s$ CNN kernels of size (1, 1) serves as both an attentive fusion layer and a dimension reducer, returning the concatenated dimensions back to their original size. Hence, the $\textbf{H}_{MA}^{cat} \in \mathbb{R}^{3*s \times 1 \times t}$ becomes $\textbf{H}_{AF} \in \mathbb{R}^{s \times 1 \times t}$. The output of the attentive fusion layer can be described as:
\begin{equation}
    \textbf{H}_{AF} = \textrm{Conv2D}([\textbf{H}_{MA}^{0}, \textbf{H}_{MA}^{1}, \textbf{H}_{MA}^{2}], kernel\_size=(1, 1)),
\end{equation}
where $[\cdot]$ is the concatenation along the kernel dimension.}
\subsection{Temporal convolutional layer}
 
\dy{TCNs are further stacked to learn the temporal dependencies on top of the space-aware temporal patterns learned from MAAF. TCNs enhance temporal sequence learning by stacking causal convolution layers, utilizing dilated 1D CNN kernels. Bai et al. \cite{bai2018empirical} improved the sequence modeling capabilities of TCNs by introducing weight normalization, residual connections, and nonlinear activation functions. The enhanced TCN model can be expressed as:
\begin{equation}
\textbf{H}^{m} = \textrm{TCN}(\textbf{H}^{m-1}) = \Phi(\textbf{H}^{m-1} + \sum_{i=0}^{k-1}f(i) \cdot \textbf{H}^{m-1}_{\text{strd} - d \cdot i})\label{TCN},
\end{equation}
where $m$ denotes the layer index, $f(\cdot)$ represents the filter, $k$ is the kernel size, $\text{strd}$ is the stride, and $d$ is the dilation factor. $\text{strd} - d \cdot i$ indicates the direction of the past. $\Phi(\cdot)$ is the PReLU activation function.}

By stacking layers of TCNs, the temporal receptive field can be increased. The receptive field size can be calculated by:
\begin{equation}
    F(m)=F(m-1)+2\times(k-1)\times d_{m}\label{eq:RF},
\end{equation}
where $m$ is the number of the convolutional block with residual connection, $k$ is the kernel size, $d_{m}$ is the dilation of the m-th convolutional block with the residual connection. When the dilation factor increases exponentially by $2$ as the number of TCN layers increases,  the receptive field can be calculated as:
\begin{equation}
    F(m) = 1 + 2 \cdot (k - 1) \cdot \sum_{i=0}^{m-1}2^{i} 
    = 1 + 2 \cdot (k - 1) \cdot (2^{m} - 1).
\end{equation}

Because SAT, the first layer of MASA-TCN, has dilation of 2,  the receptive field of MASA-TCN can be calculated as:
\begin{equation}
    \Bar{F}(m) = F(m + 1) - k + 1 = 1 + (k - 1)(2^{m+2} -3) \label{eq: RF-MASA}.
\end{equation}

\subsection{Output layer for regression and classification tasks}
Given the learned temporal representation, $\textbf{H}^{m}$, a linear layer is utilized to project it to the desired output for regression and classification, respectively. $\textbf{H}^{m}$ is a sequence of learned embeddings of the sub-segments. Since regression tasks involve n-to-n mapping, a linear layer projects the embedding of each sub-segment into a scalar, which represents the predicted emotional value of that sub-segment. Hence, the regression output, $\textbf{y}_{CER}$, can be calculated as follows:
\begin{equation}
    \textbf{y}_{CER} = [\textrm{LP}(\textbf{H}^{m}_{0}), ..., \textrm{LP}(\textbf{H}^{m}_{t-1})]  \label{eq: out_reg},
\end{equation}
where $\textrm{LP}(\cdot)$ represents the linear projection layer, and $t$ is the number of sub-segments in a sample. 

For classification tasks, the entire sequence of sub-segments corresponds to a single label. A mean fusion is applied to aggregate the predictions from all the sub-segments. Hence, the classification output, $y_{DEC}$, can be calculated as follows:
\begin{equation}
    y_{DEC} = \frac{1}{t}\sum([\textrm{LP}(\textbf{H}^{m}_{0}), ..., \textrm{LP}(\textbf{H}^{m}_{t-1}]))  \label{eq: out_cls}.
\end{equation}

\section{Experiments}
\subsection{Datasets}
Two publicly available datasets are utilized in this paper: MAHNOB-HCI \cite{5975141} for CER and DEAP \cite{5871728} for DEC.

MAHNOB-HCI\footnote{https://mahnob-db.eu/hci-tagging/} is a multimodal dataset to study human emotional responses and the implicit tagging of emotions. 30 subjects participated in the data collection experiments. Each subject watched 20 film clips, during which the synchronized recording of multi-angle facial videos, audio signals, EOG, EEG, respiration amplitude, and skin temperature were recorded. A subset \cite{7112127} of the MAHNOB-HCI database that contains 24 participants' 239 trials and the continuous labels in valence from several experts was utilized for the CER task. The final labels were determined by taking the averages of the experts' annotations. The EEG signals have 32 electrodes and the sampling rate is 256 Hz. The annotations are of 4 Hz resolution.

DEAP\footnote{http://www.eecs.qmul.ac.uk/mmv/datasets/deap/index.html} is a multimodal dataset studying human affective states. 32 subjects participated in the experiments. Each of them watched 40 1-min-long music videos while their EEG, facial expressions, and galvanic skin response (GSR) were recorded simultaneously. Self-assessments on arousal, valence, dominance, and liking from the subjects were utilized as the labels. A continuous 9-point scale was adopted to measure the levels of those dimensions, which was projected into low and high classes using a threshold of 5. The valence dimension was utilized in DEC task to be consistent with CER task. The EEG signals have 32 channels and the sampling rate is 512 Hz. 

\subsection{Preprocessing}
\label{sec_preproces}
We follow the pre-processing steps \cite{ZHANG2022108833} for MAHNOB-HCI. For each EEG trial, the first and last 30 seconds of non-stimuli durations are removed, after which an average reference is conducted. The entire trial is split into shorter segments using a 2s' sliding window with 0.25s' overlap. Then the average rPSD of (0.3-5Hz), (5-8Hz), (8-12Hz), (12-18Hz), (18-30Hz), and (30-45Hz) is calculated using Welch's method. By doing so, the 32 × 6 = 192-D rPSD features which have a frequency of 4Hz can be synchronized with the continuous labels. When training the neural networks, another sliding window whose length and step are 96 and 32 is applied to get the temporal sequence of the calculated rPSD vectors as described in \cite{ZHANG2022108833}. Hence, the size of the input to the neural networks is (batch, 192, 96). 

For DEAP, we follow \cite{9762054} to do the same pre-processing steps. For each trial of EEG, the first 3s' baseline is removed. The data is downsampled to 128 Hz. EOG artifacts were removed following the method described in \cite{5871728}. A band-pass filter from 4Hz to 45Hz is applied to remove low and high-frequency noise. Average reference is then conducted. Because MASA-TCN is designed for regression, it needs to learn from a temporal sequence of rPSD vectors. Each EEG trial is split into segments of 8 seconds, with a 4-second overlap, to create a temporal sequence for applying MASA-TCN to the DEC task and for comparison with SOTA DEC methods that utilize shorter EEG segments as input. Then the longer segments are further split into 2s' shorter segments with 0.25s' overlap to get the rPSDs in (4-8Hz), (8-12Hz), (12-18Hz), (18-30Hz), and (30-45Hz) five frequency bands. Note that the segment length in \cite{9762054} is 4 seconds; for a fair comparison, we rerun all the compared methods using 8s' segments with 4s' overlap. 

\subsection{Evaluation metrics}
The evaluation metrics for CER are the same as those in \cite{ZHANG2022108833}: root mean square error (RMSE), Pearson’s correlation coefficient (PCC), and concordance correlation coefficient (CCC). Given the prediction $\hat{\textbf{y}}$, and the continuous label $\textbf{y}$, RMSE, PCC, and CCC can be calculated as:
\begin{equation}
    RMSE = \left\| \frac{\hat{\textbf{y}}-\textbf{y}}{N} \right\|^{2} = \sqrt{\frac{\sum_{N-1}^{i=0}(\hat{y}_{i}-y_{i})^{2}}{N}},
\end{equation}

\begin{equation}
    PCC = \frac{\sigma_{\hat{\textbf{y}}\textbf{y}}}{\sigma_{\hat{\textbf{y}}}\sigma_{\textbf{y}}}=\frac{\sum_{N-1}^{i=0}(\hat{y}_{i}-\mu_{\hat{\textbf{y}}})(y_{i}-\mu_{\textbf{y}})}{\sqrt{\sum_{N-1}^{i=0}(\hat{y}_{i}-\mu_{\hat{\textbf{y}}})^{2}}\sqrt{\sum_{N-1}^{i=0}(y_{i}-\mu_{\textbf{y}})^{2}}},
\end{equation}

\begin{equation}
    CCC = \frac{2\sigma_{\hat{\textbf{y}}\textbf{y}}}{\sigma_{\hat{\textbf{y}}}^{2} + \sigma_{\textbf{y}}^{2} + (\mu_{\hat{\textbf{y}}} - \mu_{\textbf{y}})},
\end{equation}
where $N$ is the number of elements in the prediction/label vector, $\sigma_{\hat{\textbf{y}}\textbf{y}}$ is the covariance, $\sigma_{\hat{\textbf{y}}}$ and $\sigma_{\textbf{y}}$ are the variances, and $\mu_{\hat{\textbf{y}}}$ and $\mu_{\textbf{y}}$ are the means.

The evaluation metrics for DEC are the same as described in \cite{9762054}: accuracy (ACC) and F1 score. 
\subsection{Experiment settings}
\dy{For CER tasks, a leave one subject out (LOSO) strategy is utilized as described in \cite{ZHANG2022108833}. In the LOSO strategy, one subject's data is selected as test data, while the remaining subjects' data serve as training data. Within the training data, 80\% is randomly selected as training data, and the rest 20\% is utilized as validation data. We repeat this process until each subject has been the test subject once. The purposes of including an additional validation set are: 1) to provide criteria (best CCC on the validation/development set) for model and hyper-parameter selection; 2) to evaluate the model's generalization ability by testing it on unseen subject data, which is assessed only once. The mean RMES, PCC, and CCC are reported as the final results.}

\dy{We adhere to the settings described in \cite{9762054} for DEC tasks, employing a trial-wise 10-fold cross-validation for subject-specific experiments. DEC, under a generalized setting, remains challenging even in the context of subject-specific experiments \cite{9762054}. Emotion is a component of a continuous cognitive process, wherein adjacent segments within each trial demonstrate high correlations. Randomly shuffling these segments across different trials before the training-test split can lead to test data leakage, as the highly correlated adjacent segments within each trial might be present in both the training and test sets. To address such data leakage issues and adopt a more generalized evaluation methodology, we utilize trial-wise randomization to divide each subject's trials into ten folds, following the procedure outlined in \cite{9762054}. In each iteration of the 10-fold cross-validation process, one fold is allocated as the test dataset, while the remaining nine folds are divided into training and validation datasets at an 80\% and 20\% ratio, respectively. The final results are presented as the mean accuracy (ACC) and F1 score across all subjects.}

\subsection{Implementation details}
For the CER task, we follow the same training strategy as described in \cite{ZHANG2022108833}.
CCC loss is utilized to guide the training. 
\begin{equation}
    \mathcal{L}_{CCC}(\hat{\textbf{y}},\textbf{y}) = 1 - CCC(\hat{\textbf{y}}, \textbf{y}).
\end{equation}

The network is trained using the Adam optimizer, with an initial learning rate of 1e-4 and a weight decay of 1e-4. A ReduceLROnPlateau learning rate scheduler, with a patience of 5 and a reduction factor of 0.5, is also used. The maximum training epoch is set to 15 and the early stopping patient is set to 10. The batch size is set to 2. The kernel size of MASA-TCN is set to [3, 5, 15]. We tune the depth and width of MASA-TCN based on the overall performance on validation data. When the depth is 2 and the width is 64, MASA-TCN gives the best results on validation data. The dropout rate is set as 0.15 for TCNs and 0.4 for RNNs (RNN, LSTM and GRU) as suggested in \cite{7112127}. For baseline methods, we use the same training strategy and parameters as the ones of MASA-TCN for fair comparison. We also compare our results with the ones reported in the existing literatures for the same task. 

For the DEC task, based on the training strategy described in \cite{9762054}, we further reduce the maximum training epochs from 500 to 100 and add early stopping with the patient being 10 to avoid over-fitting. Besides, a two-stage training strategy is adopted. It contains two stages. In stage I, we train the model using training data and evaluate it on the validation data. The model with the best validation ACC is saved. In stage II, we combine the training and validation data as new training data and re-train the saved model on the combined dataset for at maximum 50 epochs and stop training when the training loss reaches the stopping criteria. During the training stage, the training loss of the epoch with best validation ACC is saved as the stopping criteria in the second stage. The learning rate is 1e-3 and the batch size is 32. The dropout of TCN is still 0.15 because there is a dropout operation in every TCN layer. However, this is too small for the baseline methods. Hence, the dropout rate of baseline methods is still 0.5 which is suggested in \cite{9762054}. Label smoothing with a factor of 0.1 is added to further overcome the overfitting problems. The depth and width of MASA-TCN are set to 3, and 16 based on the performance on validation data. All the baselines are re-trained using the same training strategy and the same segment length of data as MASA-TCN for a fair comparison.

\section{Results and Analysis}
\dy{In this section, we first report the CER results of MASA-TCN against several baselines, as well as the SOTA results reported in recently published papers \cite{ZHANG2022108833, 10340644}. Then the ablation study results are presented to analyze the contribution of each functional component of MASA-TCN. After that, five types of analysis experiments are conducted to analyze the effects of 1) the starting dilation, 2) the kernel size, 3) the model depth and width, 4) different fusion strategies in MAAF, and 5) early and late spatial learning. Lastly, the results for DEC tasks and the effect of mean fusion in last fully-connected layer are reported. }
\subsection{CER results on MAHNOB-HCI}
\dy{We first compare the proposed MASA-TCN with several temporal learning neural networks; then, we compare the CER results of MASA-TCN with the SOTA results reported in the existing literature \cite{ZHANG2022108833,10340644} that use the same experimental settings. Table \ref{Tab:result_loso} shows the CER results of RNN, LSTM, GRU, TCN, Chen et al.\cite{9963543}, and MASA-TCN under the LOSO experimental setting. Table \ref{Tab:result_cer_literature} lists the reported SOTA results alongside those of MASA-TCN. }

\dy{Table \ref{Tab:result_loso} shows that MASA-TCN outperforms all compared methods in RMSE, PCC, and CCC on both validation and test sets. Specifically, MASA-TCN exhibits a 14.29\% lower RMSE (an absolute drop of 0.01), a 0.043 higher PCC, and a 0.046 higher CCC than TCN. Against the RNN family, it achieves a 10.45\% lower RMSE (an absolute drop of 0.007), a 0.015 higher PCC, and a 0.031 higher CCC than the best-performing LSTM. Comparatively, MASA-TCN also outperforms SOTA results in \cite{ZHANG2022108833} with a 9.09\% lower RMSE (an absolute drop of 0.006), a 0.033 higher PCC, and a 0.04 higher CCC. Furthermore, it significantly surpasses Soleymani's methods detailed in \cite{ZHANG2022108833}, with a 25.93\% lower RMSE (an absolute drop of 0.021), a 0.08 higher PCC, and a 0.111 higher CCC. Although MASA-TCN has a slightly higher RMSE than the one in \cite{10340644}, it improves PCC and CCC by a large margin (0.037 for PCC and 0.021 for CCC). }

\begin{table*}[htp] \centering\arraybackslash
\caption{\dy{CER Results of LOSO on MAHNOB-HCI. The best results are highlighted in bold and the next best are marked using underlines.}}
\label{Tab:result_loso}
\begin{adjustbox}{center}
\begin{tabular}{lwc{6em}wc{6em}wc{6em}wc{6em}wc{6em}wc{6em}wc{6em}}
\toprule
\multirow{2}{*}{Method} & \multicolumn{3}{c}{\dy{Validation}} & \multicolumn{3}{c}{Test}                 \\ \cmidrule(lr){2-4} \cmidrule(lr){5-7}
                & \dy{RMSE$\downarrow$} & \dy{PCC$\uparrow$} & \dy{CCC$\uparrow$} & RMSE$\downarrow$& PCC$\uparrow$& CCC$\uparrow$ \\ \midrule
		     RNN &\dy{0.069$\pm$0.007} &\dy{0.493$\pm$0.114} & \dy{0.482$\pm$0.114} & 0.072$\pm$0.028 & 0.460$\pm$0.246 & 0.360$\pm$0.230\\ 
              LSTM &\dy{0.069$\pm$0.008} &\dy{0.485$\pm$0.120} &\dy{0.467$\pm$0.126}  &\underline{0.067}$\pm$0.028 & \underline{0.492}$\pm$0.242 & \underline{0.386}$\pm$0.245\\ 
              GRU & \dy{0.074$\pm$0.009}& \dy{0.465$\pm$0.126}& \dy{0.449$\pm$0.129}& 0.071$\pm$0.033 & 0.470$\pm$0.236 & 0.382$\pm$0.238\\
              TCN &\dy{\underline{0.066}$\pm$0.008} & \dy{\underline{0.505}$\pm$0.121}& \dy{\underline{0.493}$\pm$0.122} & 0.070$\pm$0.027 & 0.464$\pm$0.246 & 0.371$\pm$0.262\\
              \dy{Chen et al.} &\dy{0.108$\pm$0.048} & \dy{0.428$\pm$0.149}& \dy{0.369$\pm$0.143} & \dy{0.095$\pm$0.064} & \dy{0.475$\pm$0.230} & \dy{0.332$\pm$0.248}\\
              \textbf{MASA-TCN (ours)} &\dy{\textbf{0.060}$\pm$0.005} &\dy{\textbf{0.567}$\pm$0.090} & \dy{\textbf{0.545}$\pm$0.099}& \textbf{0.060}$\pm$0.023 & \textbf{0.507}$\pm$0.219 & \textbf{0.417}$\pm$0.236\\
              \bottomrule
		\end{tabular}
   
		\end{adjustbox}
		\begin{tablenotes}
                  \small
                  \item $\downarrow$: the lower the better; $\uparrow$: the higher the better.
            \end{tablenotes}
	\end{table*}

\begin{table}[ht] \centering\arraybackslash
\caption{\dy{Comparison with the results reported in the existing literatures \cite{ZHANG2022108833} and \cite{10340644} using the same experiment setting for CER on MAHNOB-HCI. The best results are highlighted in bold and the next best are marked using underlines.}}
\label{Tab:result_cer_literature}
\begin{adjustbox}{center}
\begin{tabular}{lwc{5.0em}wc{5.0em}wc{5.0em}}
\toprule
 Method& RMSE $\downarrow$& PCC $\uparrow$ & CCC $\uparrow$ \\
\midrule 
Soleymani et. al. \cite{7112127} & 0.081$\pm$0.034 & 0.427$\pm$0.267 & 0.306$\pm$0.257\\ 
Zhang et. al. \cite{ZHANG2022108833} & 0.066$\pm$0.025 & \underline{0.474}$\pm$0.267 & 0.377$\pm$0.250\\
\dy{Ding et. al. \cite{10340644}} & \dy{\textbf{0.059}$\pm$0.027} & \dy{0.470$\pm$0.222} & \dy{\underline{0.396}$\pm$0.225}\\ 
\textbf{MASA-TCN}  & \underline{0.060}$\pm$0.023 & \textbf{0.507}$\pm$0.219 & \textbf{0.417}$\pm$0.236\\
\bottomrule
\end{tabular}
\end{adjustbox}
\begin{tablenotes}
      \small
      \item $\downarrow$: the lower the better; $\uparrow$: the higher the better.
    \end{tablenotes}
\end{table}

\subsection{Ablation studies}
Several ablation studies are conducted to understand how each component of MASA-TCN contributes to the improvements of CER results. Starting from the baseline TCN, SAT and MAAF are gradually added to see the effects of them. The results are shown in Table \ref{Tab:result_ablation}. 

According to Table \ref{Tab:result_ablation}, adding SAT and MAAF can incrementally improve all the three metrics. By adding SAT alone, the performances of TCN can be improved by \dy{0.008} on RMSE, 0.022 on PCC, and 0.023 on CCC. The results are further improved from 0.062 to 0.060 on RMSE, from 0.486 to 0.507 on PCC, and from 0.394 to 0.417 when MAAF is also added. The results indicate the effectiveness of all those functional blocks in MASA-TCN. 

\begin{table}[ht] \centering\arraybackslash
\caption{Ablation study results of MASA-TCN on MAHNOB-HCI.}
\label{Tab:result_ablation}
\begin{adjustbox}{center}
\begin{tabular}{lwc{5.0em}wc{5.0em}wc{5.0em}}
\toprule
 Method& RMSE $\downarrow$& PCC $\uparrow$ & CCC $\uparrow$ \\
\midrule 
TCN  & 0.070$\pm$0.027 & 0.464$\pm$0.246 & 0.371$\pm$0.262\\
TCN+SA & 0.062$\pm$0.023 & 0.486$\pm$0.214 & 0.394$\pm$0.230\\
\textbf{TCN+SA+MA}  & \textbf{0.060}$\pm$0.023 & \textbf{0.507}$\pm$0.219 & \textbf{0.417}$\pm$0.236\\
\bottomrule
\end{tabular}
\end{adjustbox}
\begin{tablenotes}
      \small
      \item SA: Space-aware temporal convolutional layer; MA: Multi-anchor attentive fusion block. 
      \item $\downarrow$: the lower the better; $\uparrow$: the higher the better.
    \end{tablenotes}
\end{table}

\subsection{Effect of the starting dilation}
In this section, the effects of the starting dilation in the SAT are discussed. As mentioned in Section 3.2, the dilation of SAT starts from 2 instead of 1 which is used in TCN. There are two reasons. First, the EEG sub-segments are overlapped to synchronize them with the continuous labels, leading to some redundant information in the adjacent sub-segments. Using higher dilation at the SAT can learn the discriminative patterns effectively. Second, higher starting dilation in SAT can increase the receptive field which can reduce the number of TCN layers to get the same temporal receptive field, resulting in a more compact model size. To evaluate those effects, the starting dilation of SAT in MASA-TCN is set to 1, 2, and 4. The results are shown in Table \ref{Tab:result_first_dilation}.

\begin{table}[tp] \centering\arraybackslash
\caption{Effect of first-layer dilation of MASA-TCN on MAHNOB-HCI.}
\label{Tab:result_first_dilation}
\begin{adjustbox}{center}
\begin{tabular}{lwc{5.0em}wc{5.0em}wc{5.0em}}
\toprule
 Dilation& RMSE $\downarrow$& PCC $\uparrow$ & CCC $\uparrow$ \\
\midrule 
1 & 0.063$\pm$0.025 & 0.494$\pm$0.224 & 0.401$\pm$0.237\\ 
\textbf{2}  & \textbf{0.060}$\pm$0.023 & \textbf{0.507}$\pm$0.219 & \textbf{0.417}$\pm$0.236\\
4 & 0.062$\pm$0.022 & 0.499$\pm$0.220 & 0.411$\pm$0.243\\
\bottomrule
\end{tabular}
\end{adjustbox}
\begin{tablenotes}
      \small
      \item $\downarrow$: the lower the better; $\uparrow$: the higher the better.
    \end{tablenotes}
\end{table}

The results show that increasing the dilation to a certain degree can improve the performance, and further increase of dilation can not provide gains on the CER results. When the dilation of SAT is 2, MASA-TCN has the best performance on both validation and test set. When the value is increased further to 4, the results slightly drop on both validation and test set. The possible reasons are a dilation of 2 and a TCN layer of 2 can give enough temporal receptive field and increase more can lose certain information among the adjacent sub-segments.

\subsection{Effect of the kernel size}\label{sec_kernel_size}
\dy{This section explores the impact of varying the maximum kernel size within the MAAF, adjusting it from 3 to 15 in increments of 2. The findings, presented in Table~\ref{Tab:result_kernel_size}, reveal minimal differences in overall performance as measured by RMSE. However, larger kernel sizes are associated with improvements in both PCC and CCC, indicating that they may enhance performance in these specific metrics.}

\begin{table}[tp] \centering\arraybackslash
\caption{\dy{Effect of maximum kernel size of MASA-TCN on MAHNOB-HCI.}}
\label{Tab:result_kernel_size}
\begin{adjustbox}{center}
\begin{tabular}{lwc{5.0em}wc{5.0em}wc{5.0em}}
\toprule
 \dy{Kernel size}& \dy{RMSE $\downarrow$}& \dy{PCC $\uparrow$} & \dy{CCC $\uparrow$} \\
\midrule 
\dy{7} & \dy{0.062$\pm$0.022} & \dy{0.498$\pm$0.219} & \dy{0.404$\pm$0.239}\\ 
\dy{9} & \dy{0.061$\pm$0.021} & \dy{0.490$\pm$0.240} & \dy{0.406$\pm$0.251}\\ 
\dy{11} & \dy{0.062$\pm$0.023} & \dy{0.474$\pm$0.245} & \dy{0.393$\pm$0.244}\\ 
\dy{13} & \dy{\textbf{0.060}$\pm$0.022} & \dy{0.503$\pm$0.216} & \dy{0.411$\pm$0.244}\\ 
\dy{15} & \dy{0.060$\pm$0.023} & \dy{\textbf{0.507}$\pm$0.219} & \dy{\textbf{0.417}$\pm$0.236}\\ 
\bottomrule
\end{tabular}
\end{adjustbox}
\begin{tablenotes}
      \small
      \item \dy{$\downarrow$: the lower the better; $\uparrow$: the higher the better.}
    \end{tablenotes}
\end{table}

\begin{figure}[t]
\centering
    \subfigure[Effect of the depth]{
    \includegraphics[width=0.46\linewidth]{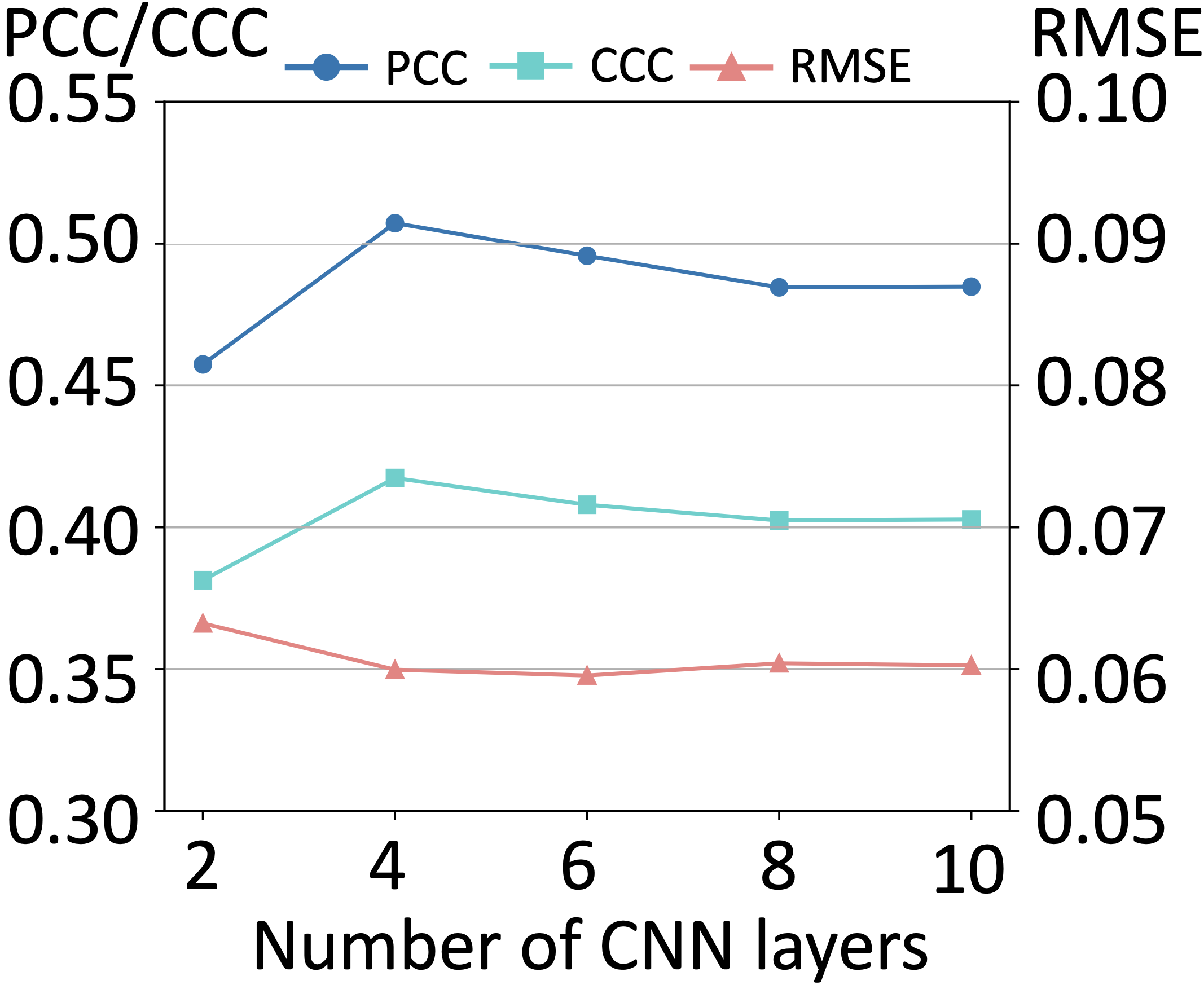}
    }
    \subfigure[Effect of the width]{
    \includegraphics[width=0.46\linewidth]{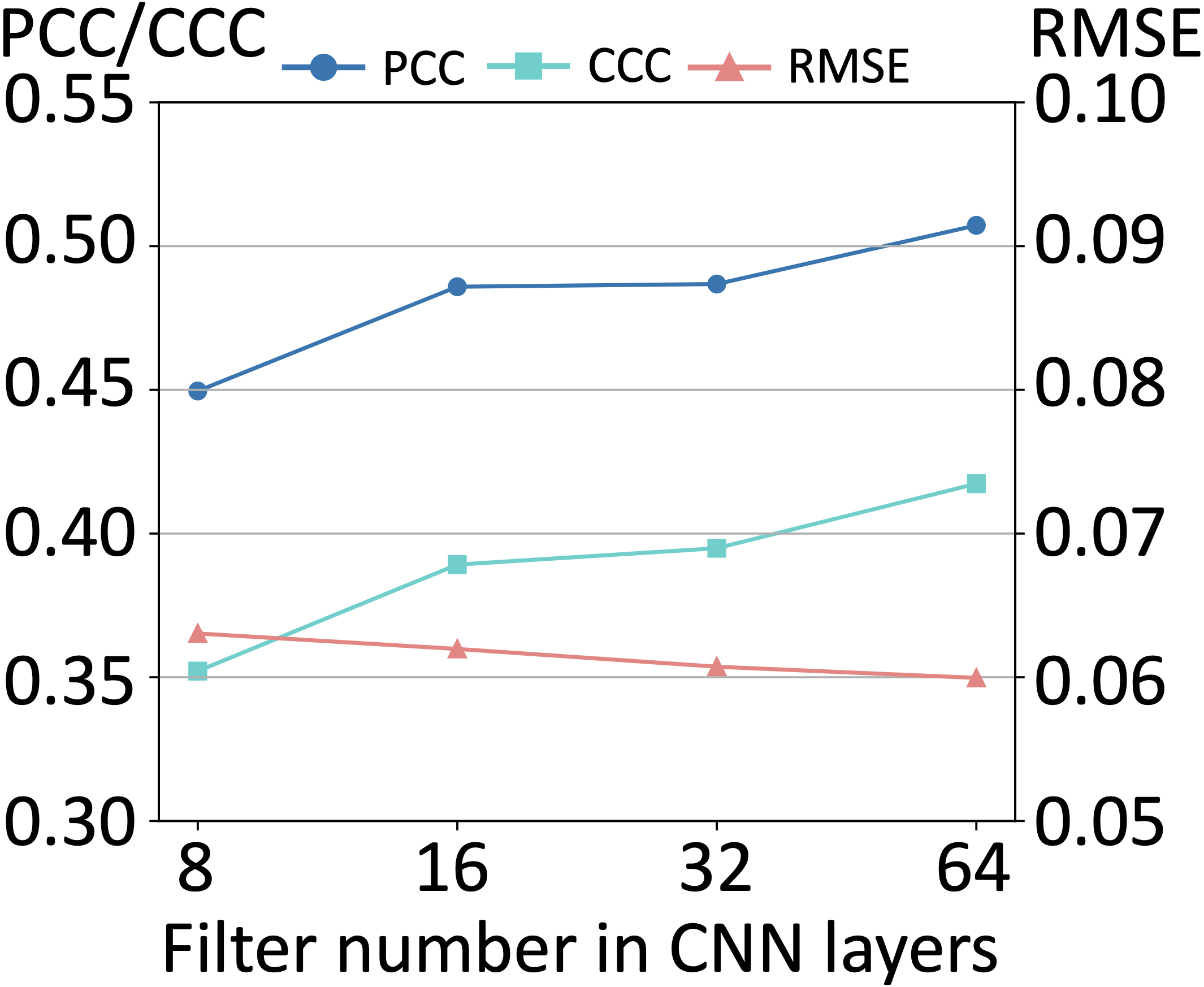}
    }
\caption{Effect of the depth and width of MASA-TCN.}
\label{fig:depth-and-width}
\end{figure}

\subsection{Effect of the model depth and width}
Experiments about the effects of model depth and width are conducted to better understand MASA-TCN. For model depth studies, the SAT is regarded as 2 layers due to the sequential operation of two types of CNN kernels. And there are 2 causal convolutional layers in one TCN layer. Hence the depths are set as 2, 4, 6, 8, and 10. For the width, it is the number of kernels in each CNN layer. The widths are set as 8, 16, 32, and 64. The results are shown in Fig. \ref{fig:depth-and-width}.

According to the results, depth is not sensitive when it is higher than 4, and the width more sensitively affects the model performance compared with the depth. From Fig. \ref{fig:depth-and-width} (a), only having SAT and MAAF can not provide good performance. This is because the temporal receptive field is not enough. When the depth is 4, MASA-TCN achieves the best performance on both validation and test sets. However, when the depth increases to higher than 4, the performances drop a little bit and become stable. This is due to that enough temporal receptive field is achieved and a deeper model is relatively harder to train than the shallow one \cite{7780459}. From Fig. \ref{fig:depth-and-width} (b), the performances are positively related to the width. And when the width is 64, MASA-TCN achieves the best results on both validation and test sets. Note that we also conducted an experiment that use 128 as the width, but the model gave very low performances, which indicates the wider model is also harder to train. 

\subsection{Effect of different fusion strategies in MAAF}
The effects of different fusion strategies in MAAF are analyzed and discussed in this section. Because three SATs with different kernel lengths are parallelly utilized in MAAF, the output needs to be fused for the subsequent TCNs. Three types of fusion mechanisms are studied: concatenation, mean, and attentive fusion.

Based on the results in Table \ref{Tab:result_fusion}, all three types of fusion methods achieve relatively acceptable performances, and with attentive fusion, MASA-TCN has the best performances on both validation and test sets. This indicates the effectiveness of attention fusion in MAAF. 

\begin{table}[ht] \centering\arraybackslash
\caption{Effect of fusion strategy in MAAF on MAHNOB-HCI.}
\label{Tab:result_fusion}
\begin{adjustbox}{center}
\begin{tabular}{lwc{5.0em}wc{5.0em}wc{5.0em}}
\toprule
 Fusion Method& RMSE $\downarrow$& PCC $\uparrow$ & CCC $\uparrow$ \\
\midrule 
Concatenate & 0.062$\pm$0.023 & 0.476$\pm$0.239 & 0.395$\pm$0.249\\ 
Mean  & 0.062$\pm$0.025 & 0.496$\pm$0.234 & 0.410$\pm$0.255\\ 
\textbf{Attentive}  & \textbf{0.060}$\pm$0.023 & \textbf{0.507}$\pm$0.219 & \textbf{0.417}$\pm$0.236\\
\bottomrule
\end{tabular}
\end{adjustbox}
\begin{tablenotes}
      \small
      \item $\downarrow$: the lower the better; $\uparrow$: the higher the better.
    \end{tablenotes}
\end{table}

\subsection{Effect of early and late spatial learning}
The order of spatial learning is studied in this part. As illustrated in Section 2.2, there are spatial, spectral, and temporal patterns that need to be recognized for EEG data. Typically the spatial patterns can be learned by a 1D CNN kernel whose size is $(c, 1)$, where $c$ is the number of EEG channels. In MASA-TCN, the spatial learning is done in SAT, which is regarded as early spatial learning. The spatial patterns can also be learned after the last several TCN layers, which is termed late spatial learning. We compare both early and late spatial learning. The results are shown in Table \ref{Tab:result_spatial}.

Early spatial learning is more effective than late spatial learning according to the results. It is noticeable that late spatial learning cannot even has comparable performance with the one using early spatial learning. More analyses should be done in the future to better understand the reason. 

\begin{table}[tp] \centering\arraybackslash
\caption{Effect of early and late spatial learning of MASA-TCN on MAHNOB-HCI.}
\label{Tab:result_spatial}
\begin{adjustbox}{center}
\begin{tabular}{lwc{5.0em}wc{5.0em}wc{5.0em}}
\toprule
 Spatial Learning& RMSE $\downarrow$& PCC $\uparrow$ & CCC $\uparrow$ \\
\midrule 
Late & 0.434$\pm$0.602 & 0.153$\pm$0.250 & 0.115$\pm$0.189\\ 
\textbf{Early}  & \textbf{0.060}$\pm$0.023 & \textbf{0.507}$\pm$0.219 & \textbf{0.417}$\pm$0.236\\
\bottomrule
\end{tabular}
\end{adjustbox}
\begin{tablenotes}
      \small
      \item $\downarrow$: the lower the better; $\uparrow$: the higher the better.
    \end{tablenotes}
\end{table}

\subsection{DEC results on DEAP}
\subsubsection{Comparison with baselines for DEC}
MASA-TCN achieves SOTA performances on CER tasks, we further explore the possibility of extending it to DEC tasks and compare it with the SOTA methods of DEC tasks, SVM (2012) \cite{5871728},
DeepConvNet (2017) \cite{https://doi.org/10.1002/hbm.23730}, EEGNet (2018) \cite{Lawhern_2018}, TSception (2022) \cite{9762054}, \dy{and MEET (2023) \cite{10345766}} in this section. Because MASA-TCN is mainly designed for CER tasks, a regressor is utilized to generate a 1D output that has the same length as the continuous labels. One way to adapt MASA-TCN to DEC is to change the output size of the regressor from 1D to binary output and the regressor becomes a normal classifier in most deep learning methods for classification. However, we can also extend MASA-TCN to DEC by using a mean fusion on the output of the regressor as a kind of classifier ensemble which can increase the robustness. Hence, in MASA-TCN, we choose the latter to extend it from the CER tasks to the DEC tasks. Note that for a fair comparison, all the methods use the same data preprocessing steps, the same segment length (8s) with a overlapping of 50\%, and the same training strategies as the ones of MASA-TCN. The results are shown in Table \ref{Tab:result_DEC_valence} and \ref{Tab:result_DEC_arousal}.

\dy{As demonstrated in Table~\ref{Tab:result_DEC_valence} and Table~\ref{Tab:result_DEC_arousal}, the MASA-TCN model delivers superior or comparable performance in emotion classification tasks. Specifically, MASA-TCN exhibits the highest accuracy and F1 score in the valence dimension. While the differences in accuracy among various deep learning approaches are not markedly significant, MASA-TCN and MEET, the latter securing the second place, outshine their counterparts in terms of F1 scores. In the context of the arousal dimension, MASA-TCN attains the highest accuracy, whereas TSception leads with the best F1 score.}

\begin{table}[tp]
\centering
\caption{\dy{Results of DEC task for valence dimension using DEAP. The best results are highlighted in bold and the next best are marked using underlines.}}
\begin{tabular*}{0.7\linewidth}{
  @{\extracolsep{\fill}\hspace{\tabcolsep}}
  l c c 
} 
\toprule
 Method & {ACC(\%)} & {F1(\%)}\\
\midrule
SVM & 55.17$\pm$7.29 &58.07$\pm$10.73 \\ 
EEGNet & 58.45$\pm$9.30 & 58.14$\pm$13.57\\ 
DeepConvNet & 59.30$\pm$10.45  & 60.07$\pm$13.57 \\ 
TSception & 58.71$\pm$9.12 & 57.09$\pm$19.16 \\ 
\dy{MEET} & \dy{\underline{59.44}$\pm$9.45} & \dy{\underline{63.35}$\pm$12.15}\\
\textbf{MASA-TCN}  & \textbf{60.20$\pm$8.13} & \textbf{64.58$\pm$12.61}\\ 
\bottomrule
\end{tabular*}
\label{Tab:result_DEC_valence}

\end{table}

\begin{table}[tp]
\centering
\caption{\dy{Results of DEC task for arousal dimension using DEAP. The best results are highlighted in bold and the next best are marked using underlines.}}
\begin{tabular*}{0.7\linewidth}{
  @{\extracolsep{\fill}\hspace{\tabcolsep}}
  l c c 
} 
\toprule
\dy{ Method} & \dy{ACC(\%)} & \dy{F1(\%)}\\
\midrule
\dy{SVM} & \dy{60.57$\pm$11.10} &\dy{58.62$\pm$23.83} \\ 
\dy{EEGNet} & \dy{59.85$\pm$8.80} & \dy{62.16$\pm$15.21}\\ 
\dy{DeepConvNet} & \dy{59.53$\pm$9.09}  & \dy{61.48$\pm$17.10} \\ 
\dy{TSception} & \dy{\underline{61.50}$\pm$11.27} & \dy{\textbf{62.62$\pm$17.21}} \\ 
\dy{MEET} & \dy{60.37$\pm$12.08} & \dy{58.74$\pm$ 24.43}\\
\dy{\textbf{MASA-TCN}}  & \dy{\textbf{62.09$\pm$10.39}} & \dy{\underline{62.24}$\pm$19.29}\\ 
\bottomrule
\end{tabular*}
\label{Tab:result_DEC_arousal}

\end{table}

\subsubsection{Effect of mean fusion in the fully connected layer for DEC}
\dy{This section examines the impact of mean fusion in the final fully connected (FC) layer through two experiments: (1) removing mean fusion from MASA-TCN and (2) implementing mean fusion in the last FC layer of MASA-TCN. Results in Table \ref{Tab:result_DEC_fusion} reveal that incorporating mean fusion into MASA-TCN enhances the ACC and F1 score by 1.63\% and 2.7\%, respectively, highlighting the benefit of mean fusion in MASA-TCN.}

\begin{table}[t]
\centering
\caption{\dy{Effect of mean fusion in the FC of MASA-TCN for DEC on DEAP.}}
\begin{tabular*}{0.8\linewidth}{
@{\extracolsep{\fill}\hspace{\tabcolsep}}
l c c
} 
\toprule
\dy{Method} & \dy{ACC(\%)} & \dy{F1(\%)} \\
\midrule
\dy{MASA-TCN w/o MF} & \dy{58.57$\pm$7.92} & \dy{61.88$\pm$12.75} \\ 
\dy{\textbf{MASA-TCN w MF}}  & \dy{\textbf{60.20}$\pm$8.13} & \dy{\textbf{64.58}$\pm$12.61}\\ 
\bottomrule
\end{tabular*}
\label{Tab:result_DEC_fusion}
\begin{tablenotes}
      \small
      \item \dy{MF: mean fusion in the last fully connected layer.}
    \end{tablenotes}
\end{table}

\begin{figure}[tp]
\centering
    \subfigure[Well regressed]{
    \includegraphics[width=0.46\linewidth]{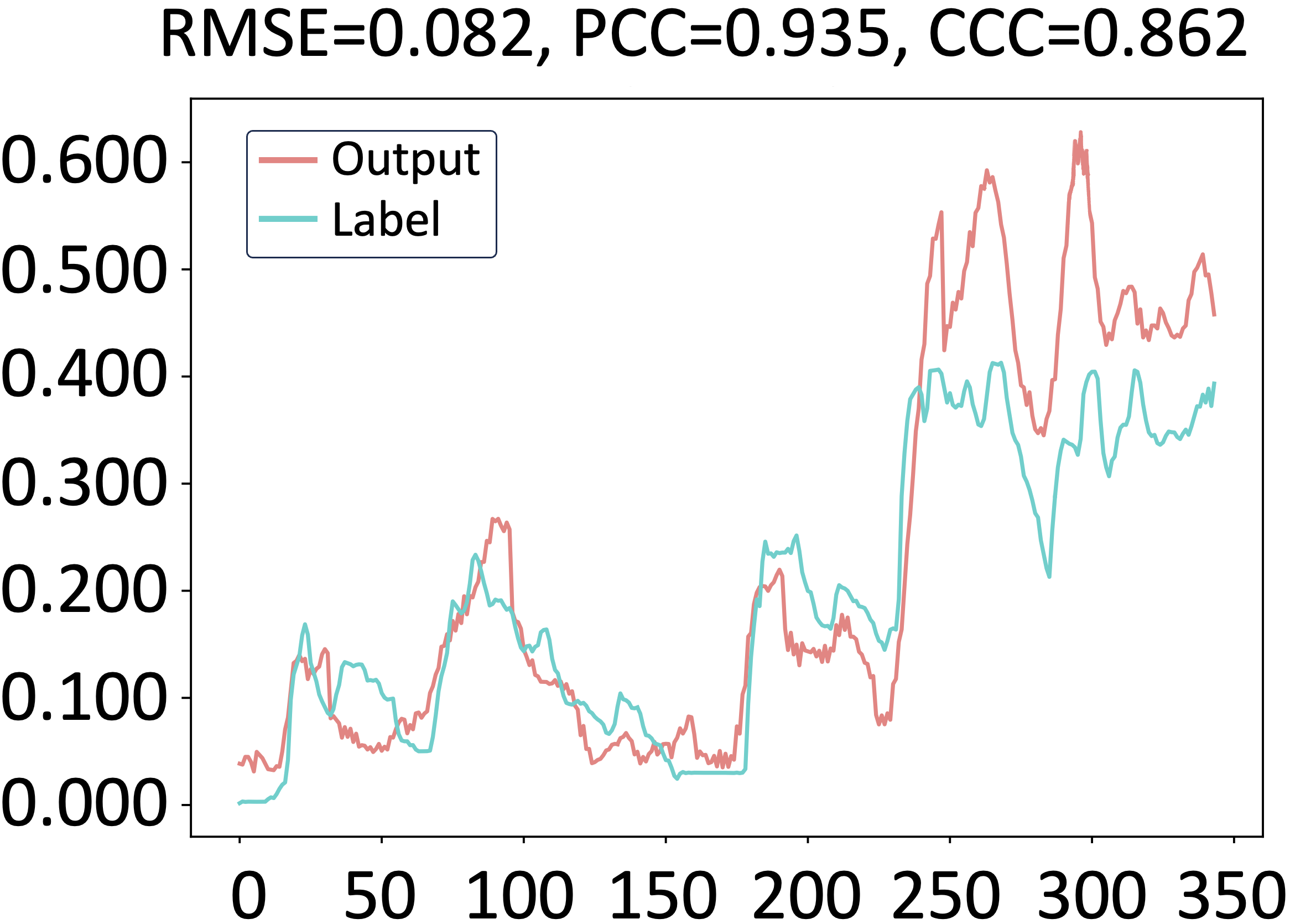}
    }
    \subfigure[Moderately regressed]{
    \includegraphics[width=0.46\linewidth]{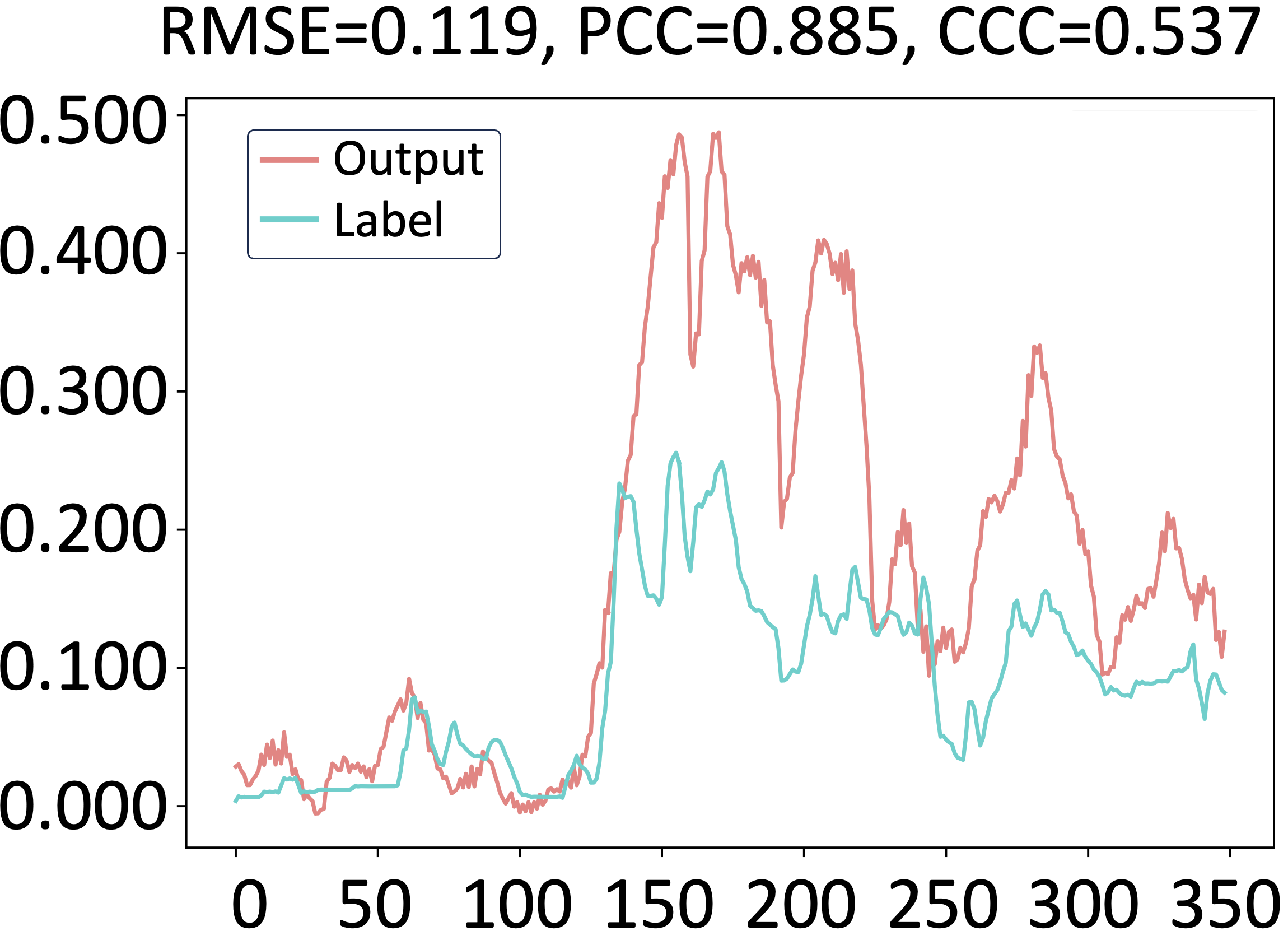}
    }
    \subfigure[Poorly regressed]{
    \includegraphics[width=0.46\linewidth]{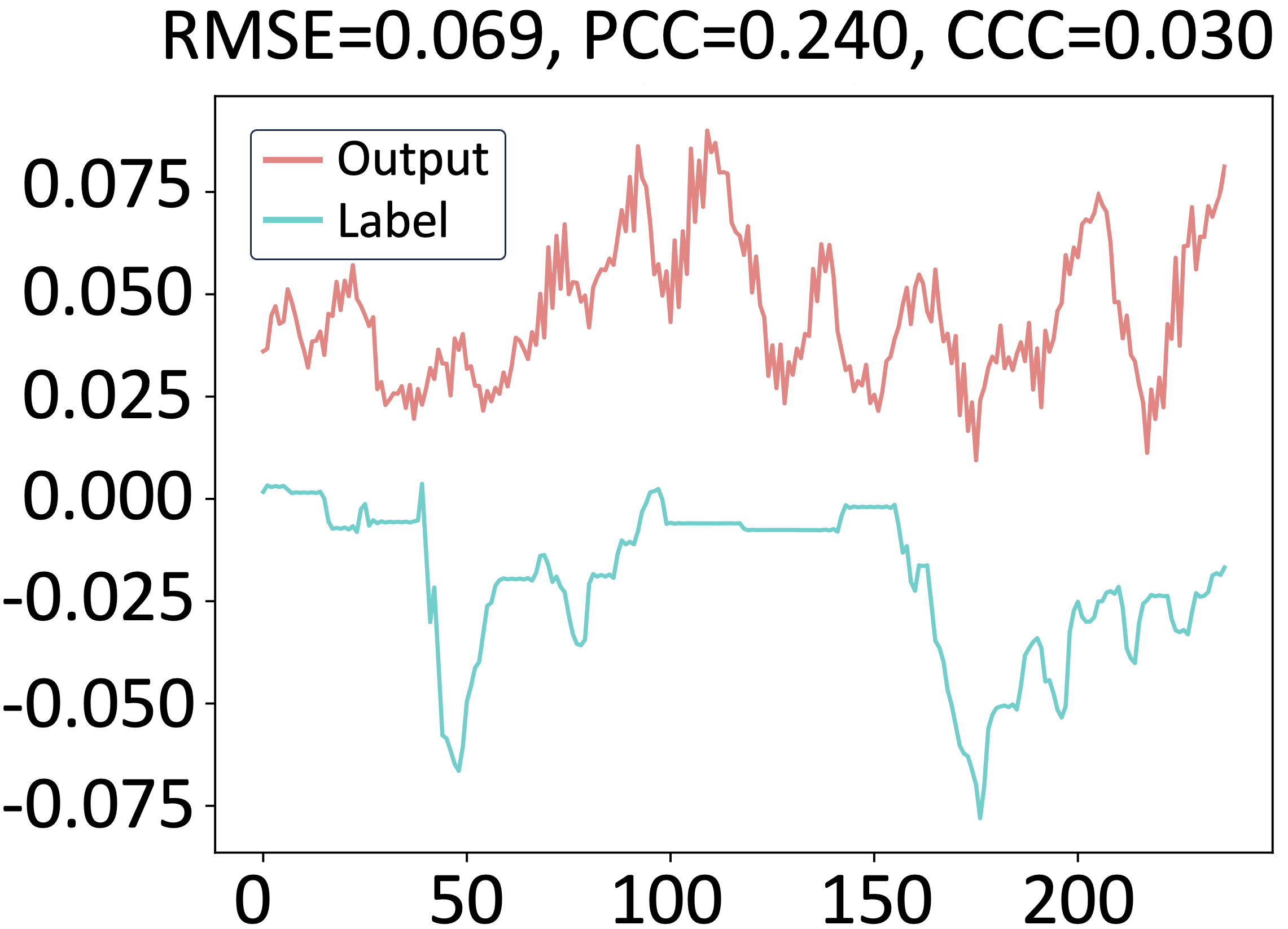}
    }
    \subfigure[Poorly regressed]{
    \includegraphics[width=0.46\linewidth]{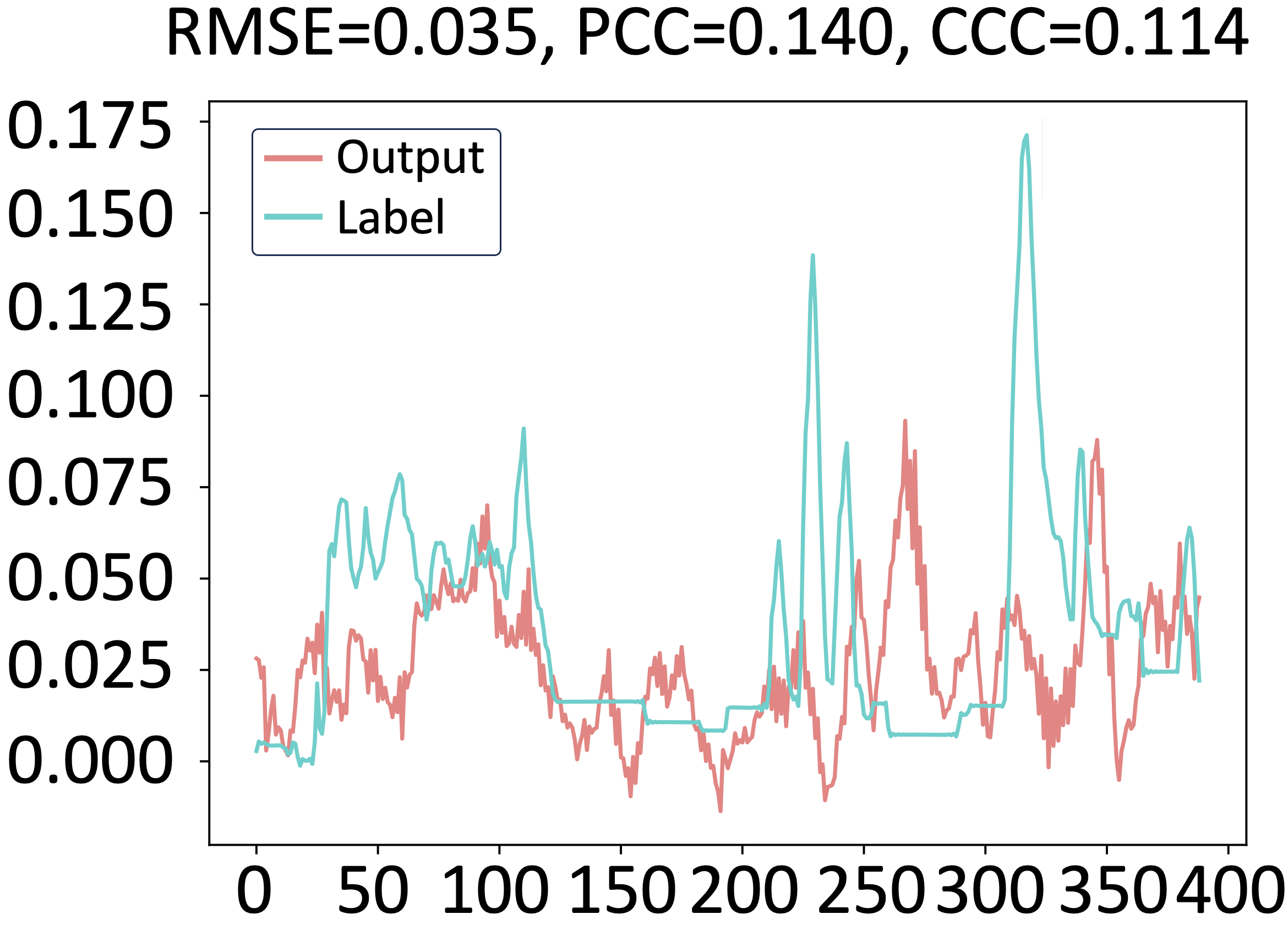}
    }
\caption{Four representative samples of well and poorly regressed trials of MASA-TCN for CER. The y-axis is the valence score, and the x-axis is the index of the samples along temporal dimension.}
\label{fig:oupt_label_visulization}
\end{figure}
\section{Discussion}
The CER tasks are relatively more comprehensive to study human emotions. CER tasks require the model to predict the temporally continuous labels of emotions using EEG signals, which are rarely explored in the existing literatures \cite{ZHANG2022108833}. Emotion is a continuous neural cognitive process of the brain \cite{doi:10.1177/1754073915590618}. In general EEG collection experiments in the studies for emotions \cite{5871728, 5975141}, the subjects are required to watch and listen to the affective stimuli for a certain duration. And the emotional states are not consistent during the entire trial \cite{9698041}. By refining the label of shorter segments using the continuous label instead of the single label of one trial in DEC, improvements in classification are observed \cite{9176111}. Despite the importance of exploring novel methods for EEG CER tasks, only a few works \cite{ZHANG2022108833, 7112127} have proposed some algorithms. And all of them use flattened feature vectors as input while not effectively learning the spatial patterns across EEG channels. 
 
MASA-TCN is proposed in this paper to enable TCN to learn spatial, spectral, and temporal patterns simultaneously for the CER tasks. The main functional block for spatial learning is the SAT layer that consists of context kernel and spatial fusion kernel two types of CNN kernels. With the help of zero padding and dilation along the temporal dimension, SAT can also learn the temporal causal dependencies. Because EEG contains abundant temporal information that is related to the brain's emotional activities changes from time to time, and the temporal dependencies might happen in different time scales \cite{9762054, doi:10.1177/1754073915590618}, a MAAF block is further designed to capture those temporal dynamics with the help of multiple temporal kernel lengths as well as an attentive fusion layer. Extensive experiments on a publicly available dataset have been done to evaluate the proposed method. The results demonstrate the effectiveness of MASA-TCN for CER tasks and we set new SOTA results against the recently published results in \cite{ZHANG2022108833}. We further extend MASA-TCN from regression tasks to classification tasks by adding a mean fusion in the final fully-connected layer (regressor). It also achieves higher classification results over several SOTA methods in DEC tasks. To the best of our knowledge, this is the first work to propose a unified model for both CER and DEC tasks. The experiment also indicates that calculating the mean of the output of a regressor as the classification output can yield a certain improvement in F1 scores.

Besides the analysis experiments we conducted and discussed in Section 5, some discussions on the output of MASA-TCN for CER are given here. Four representative samples for well, moderately, and poorly regressed trials are selected to show the differences between the prediction and ground truth. They are shown in Fig. \ref{fig:oupt_label_visulization}. The discussions are two-fold: the performance of MASA-TCN for CER and the differences among the three evaluation metrics. 
 
We first discuss the performance of MASA-TCN. According to Fig. \ref{fig:oupt_label_visulization} (a) and (b), MASA-TCN can well regress the relatively smaller absolute value ($>0.15$) while the predictions of larger-value labels, especially the ones with sudden changes, are not well addressed. In the future, some regularization terms can be added to the output of MASA-TCN to reduce the amplitude after sudden changes. It is also noticeable that MASA-TCN handles the positive labels better than the negative ones by comparing Fig. \ref{fig:oupt_label_visulization} (a), (b), and (c).  Based on Fig. \ref{fig:oupt_label_visulization} (d), it can be seen that MASA-TCN can not well regress the details of sudden short-term fluctuations. RMSE can punish the distance between prediction and label point-wise, hence, it is worth trying to guide the training using a weighted combination of RMSE and CCC for better regression of the details instead of using CCC loss only. Next, we give some discussions on the evaluation metrics. 
 
CCC is a better evaluation metric for CER compared with RMSE and PCC. RMSE focuses on point-to-point precision, while the correlation between the predictions and labels is not effectively measured. As shown in  Fig. \ref{fig:oupt_label_visulization}, when the trial is well regressed (Fig. \ref{fig:oupt_label_visulization} (a)), RMSE is still larger than the ones of the poorly regressed ones (Fig. \ref{fig:oupt_label_visulization} (c) and Fig. \ref{fig:oupt_label_visulization} (d)). That's because the labels are in relatively lower amplitudes in Fig. \ref{fig:oupt_label_visulization} (c) and (d) than the ones in Fig. \ref{fig:oupt_label_visulization} (a). Even though the trends are not well regressed, the point-to-point distances are still small. However, CCCs of those two poorly regressed outputs are much lower than the well-regressed ones because CCC also measures the correlation between the two vectors. Although PCC can measure the correlations, it ignores the absolute distances among points of the outputs and labels. Hence, we can see the PCC is still high in Fig. \ref{fig:oupt_label_visulization} (b) even though there are long drifts between the outputs and labels. CCC can reflect those drifts as well, hence, the CCC of Fig. \ref{fig:oupt_label_visulization} (a) is much higher than the one of Fig. \ref{fig:oupt_label_visulization} (b). 
 
There are also some limitations in this work that need to be discussed. The first one is the lack of datasets for CER tasks. This is a common problem for EEG CER tasks. Because preparing a dataset for the EEG CER tasks needs well-designed experiment protocols as well as the efforts of a number of experts to continuously annotate the corresponding trials. In the future, more datasets need to be created to further boost this research area. Besides, more interpretability methods should be applied to better understand why early spatial learning is much better than late spatial learning. At last, in this paper, we follow \cite{ZHANG2022108833} that only uses CCC in the loss function to guide the training process. In the future, using a weighted combination of RMSE, PCC, and CCC in the loss function is expected to provide certain improvements. 

\section{Conclusion}
In this paper, MASA-TCN is proposed to improve the SOTA results of the CER and DEC tasks using EEG. Compared with the SOTA methods \cite{ZHANG2022108833, 7112127} that don't effectively learn the spatial patterns among EEG channels, a novel SAT layer is designed to enable TCN to capture spatial, spectral, and temporal patterns simultaneously. A MAAF block is further proposed to capture the temporal dynamics that might happen in different time scales underlying emotional cognitive processes. By adding a mean fusion in the output of the regressor of MASA-TCN, we further extend MASA-TCN from CER to DEC, making it a unified model for both the CER and DEC tasks using EEG. Extensive experiments on two public emotion datasets show the effectiveness of the proposed methods for both CER and DEC. New SOTA results are achieved by MASA-TCN for those tasks. 

\section*{References}

\bibliographystyle{./IEEEtran}
\bibliography{./mybib}

\end{document}